\documentclass[letterpaper]{article} 
\usepackage[preprint]{aaai2027}  
\usepackage[hyphens]{url}  
\usepackage{graphicx} 
\usepackage{natbib}  
\usepackage{caption} 
\usepackage{algorithm}

\usepackage{newfloat}
\usepackage{listings}
\DeclareCaptionStyle{ruled}{labelfont=normalfont,labelsep=colon,strut=off} 
\floatstyle{ruled}
\newfloat{listing}{tb}{lst}{}
\floatname{listing}{Listing}

\usepackage{booktabs}

\usepackage[utf8]{inputenc}
\usepackage[T1]{fontenc}
\usepackage{amsmath,amssymb,amsfonts}
\usepackage{graphicx}
\usepackage{booktabs}
\usepackage{algpseudocode}
\usepackage{xcolor}
\usepackage{multirow}
\usepackage{subcaption}
\usepackage{cite}
\usepackage{enumitem}
\usepackage{tabularx}
\usepackage{tikz}
\usepackage{pgfplots}
\pgfplotsset{compat=1.18}
\usetikzlibrary{shapes.geometric,arrows.meta,positioning,calc,fit,backgrounds,patterns,decorations.pathreplacing}

\newcommand{\sysname}{\textsc{BEVPIPE}}

\title{The Operator Mismatch Problem: Deploying BEV Perception with Portable GPU Compute}
\author {
    Rohit Verma,
    Anand V Bodas,
}
\affiliations {
    Intel\\
    rohit1.verma@intel.com, anand.v.bodas@intel.com
}

\begin{document}

\maketitle

\begin{abstract}
Modern autonomous driving systems rely on bird's-eye-view (BEV) perception models that fuse camera and LiDAR inputs to detect objects in 3D space.  These models are accurate, but they cannot be deployed through standard inference runtimes.  The reason is an \emph{operator mismatch} between dense convolutions (which runtimes handle well), sparse 3D convolutions (which runtimes cannot represent), and geometric scatter operations (which runtimes have no vocabulary for).  Today, every sparse convolution library is CUDA-only and PyTorch-coupled, locking BEV deployment to a single vendor's hardware and a single execution framework.

We present \sysname{}, a framework for deploying multi-modal BEV perception pipeline using portable GPU compute APIs, integrated with production inference runtime. \sysname{} partitions the model into runtime-managed dense subgraphs and three external operator extensions (voxelizer, sparse encoder, BEV projector), connected through a shared GPU memory space. \sysname{} achieves a $19.5\times$ end-to-end speedup over conventional deployments while retaining $98.5\%$ of reference mAP. We also showcase that \sysname{} is portable across different GPU backends.
\end{abstract}


\section{Introduction}
\label{sec:introduction}

Autonomous vehicles perceive the world through sensors and cameras that see color and texture, and LiDAR scanners that measure precise 3D geometry.  The most accurate perception systems fuse both modalities into a unified bird's-eye view (BEV) representation, where objects are detected in the coordinate frame relevant to planning~\cite{liu2023bevfusion,liang2022bevfusion,li2022bevformer}. Although training these models is a solved problem, deploying them is not.

The difficulty lies in a mismatch between what these models compute and what inference runtimes understand.  A BEV perception pipeline interleaves three fundamentally different kinds of computation: \textbf{Dense operations} consisting of image convolutions, transformers, and fully-connected layers, that operate on regular tensors with static shapes; \textbf{Sparse 3D convolutions}, which are applied to the irregular set of occupied voxels in a point cloud, where the number of active locations and their connectivity change with every input; and \textbf{Geometric scatter operations} which include voxelization (hashing points into bins) and BEV projection (scattering depth-weighted features into a grid), that involve hash tables, atomic accumulation, and coordinate-dependent branching. 




When a runtime encounters an operator it cannot represent, it has three options: (a) fall back to a slow CPU reference implementation, (b) write a custom kernel plugin, or (c) reject the model entirely~\cite{tensorrt2024,openvino2024}.  The existence of entire libraries such as SpConv~\cite{spconv2022}, TorchSparse~\cite{tang2022torchsparse, tang2023torchsparse++}, MinkowskiEngine~\cite{choy2019minkowski}, built outside the inference stack is direct evidence that runtimes cannot handle this class of computation.\footnote{Recent TensorRT versions support dynamic shapes via \texttt{INetworkDefinition} profiles, but tensors whose \emph{index sets} change at runtime, as in sparse convolution, remain outside the scope of static graph compilation.}

This \emph{operator mismatch} has a concrete consequence: state-of-the-art BEV perception models cannot be deployed through standard inference pathways without substantial custom engineering. And that engineering, today, is locked to a single hardware vendor. Every existing sparse convolution library is CUDA-only and PyTorch-coupled~\cite{spconv2022,tang2022torchsparse,choy2019minkowski}. BEVFusion's own deployment uses TensorRT with custom CUDA plugins~\cite{liu2023bevfusion}.  If one wants to run BEV perception on Intel, AMD, or mobile GPUs, or through any runtime other than TensorRT, they are on their own.

In this paper, we answer the question: Can a complete multimodal BEV perception pipeline be deployed efficiently using portable GPU compute APIs (OpenCL/SYCL) without CUDA and integrated with a production inference runtime? We present \sysname{}, a system that does this by partitioning the pipeline between the runtime (which handles dense operations) and three portable OpenCL GPU extensions (which handle the rest). Answering this question involves addressing the below research questions.

\begin{description}[nosep,font=\normalfont\bfseries,style=unboxed]
\item[Q1.] How should a heterogeneous perception pipeline be split between a runtime's native operators and external custom kernels?
\item[Q2.] How does the granularity of that split affect overhead?
\item[Q3.] Where does execution time actually go in each operator class?
\end{description}

The key contributions of this work are summarized below which address these questions.

\begin{itemize}[leftmargin=*,nosep]

\item We formalize the operator-mismatch problem as a partition boundary placement problem and demonstrate which partitioning approach works effectively.

\item We design \sysname{}, an end-to-end BEV perception system suitable for portable GPU compute APIs and integrated with a production inference runtime, enabling deployment on Intel, AMD, and other hardwares without modification.

\item Our experiments demonstrate the effectiveness of the proposed system in terms of both performance and accuracy. We also showcase that the pipeline partitioning methodology generalizes beyond OpenCL to SYCL, enabling a single pipeline to target multiple GPU backends.

\end{itemize}

\section{Related Work}
\label{sec:related_work}

The operator mismatch problem has been addressed from several directions, including specialized sparse convolution libraries, inference runtime extensions, compiler-based code generation, and direct BEV deployment efforts.  Each addresses part of the problem but leaves a gap that motivates our work.

\subsection{Sparse Convolution Libraries}

Sparse 3D convolution~\cite{graham2015sparse} and its submanifold variant~\cite{graham2018submanifold} are implemented by three mature libraries: SpConv~\cite{spconv2022} (fused gather--GEMM--scatter CUDA kernels), TorchSparse/++~\cite{tang2022torchsparse,tang2023torchsparse++} (locality-driven reordering and auto-tuned kernel selection), and MinkowskiEngine~\cite{choy2019minkowski} (generalized sparse convolution).

However, all three are CUDA-only and PyTorch-coupled, cannot be composed with a runtime's graph optimizer, memory planner, or INT8 pipeline, and lock deployment to NVIDIA hardware.  For production on Intel, AMD, or Qualcomm GPUs, they are unavailable.

\subsection{Inference Runtimes and Compilers}

OpenVINO~\cite{openvino2024}, TensorRT~\cite{tensorrt2024}, ONNX~Runtime~\cite{onnxruntime2021}, and TVM~\cite{chen2018tvm} optimize dense models via fusion, memory planning, and hardware-targeted codegen, and all expose custom-operator extension APIs.  However, the runtime treats custom operators as opaque black boxes~\cite{chen2018tvm,zheng2020ansor}: graph fusion stops at extension boundaries and memory planning cannot span runtime- and extension-managed buffers.  Thus \emph{how} and \emph{at what granularity} we partition has first-order impact, a trade-off prior works have not analyzed.  Compiler stacks (Ansor~\cite{zheng2020ansor}, Triton~\cite{tillet2019triton}) can generate novel-operator code but do not integrate the INT8 calibration, memory pools, and accelerated dense kernels that vendor runtimes provide, forcing reimplementation of existing functionality.

\subsection{BEV Model Deployment}

The closest prior work is BEVFusion's deployments~\cite{liu2023bevfusion, liang2022bevfusion}, using custom CUDA plugins for sparse convolution and BEV projection. MMDeploy offers a similar NVIDIA-stack pipeline~\cite{mmdeploy2021}.

These show BEV deployment \emph{can} work on one runtime and one hardware family, but do not analyze the methodology such as partitioning criteria, boundary cost trade-offs, operator-heterogeneity/scheduling interaction, nor the portability question of whether the approach holds when CUDA is unavailable?

\subsection{Geometric Operations}

Lift-Splat-Shoot~\cite{philion2020lift} established depth-based camera-to-BEV projection, with efficient CUDA kernels~\cite{liu2023bevfusion}, geometry caching~\cite{huang2022bevdet4d}, and attention-based alternatives~\cite{li2022bevformer}; voxelization uses GPU hash tables with atomic accumulation~\cite{yan2018second,choy2019minkowski}.  Both involve irregular access and atomic contention, poor fits for dense runtimes \emph{regardless of vendor}.

\subsection{What a Solution Needs}

Surveying these state-of-the-art approaches, we identify three properties that a successful deployment approach must have:

\begin{enumerate}[leftmargin=*,nosep]
\item \textbf{Portability:} no single platform dependency; must work through portable GPU compute APIs (OpenCL, SYCL) that target multiple hardware vendors.
\item \textbf{Runtime integration:} sparse and geometric kernels must compose with the runtime's dense optimization pipeline, not bypass it entirely.
\item \textbf{Partition-aware design:} the boundary between runtime-native and custom operators must be explicitly managed to minimize synchronization and format conversion overhead.
\end{enumerate}

Next, we provide an overview of \sysname{}, which is designed to satisfy all three.

\section{System Overview}
\label{sec:system_overview}

In this section, we describe the BEV perception pipeline, what a deployment system must handle, and a broad overview of the \sysname{} architecture.

\begin{figure*}[!ht]
    \centering
    \includegraphics[width=0.6\linewidth]{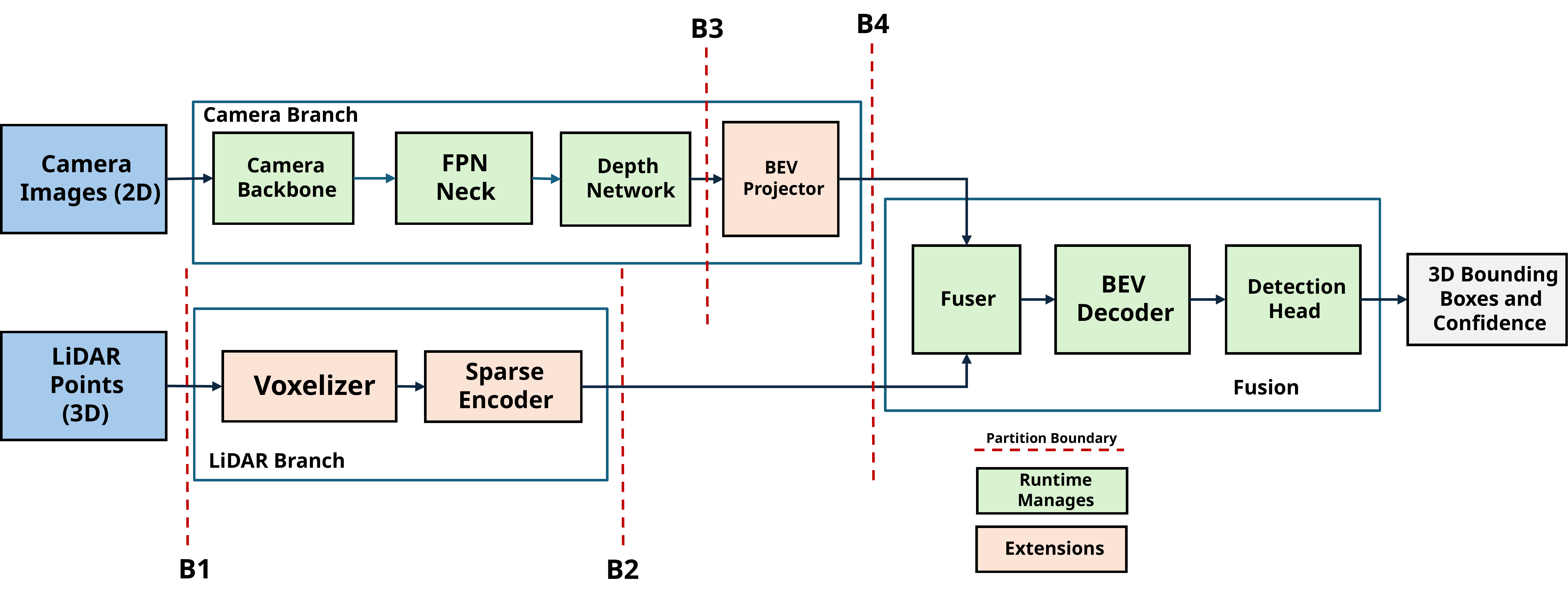}
    \caption{System architecture of \sysname{}.}
    \label{fig:system_architecture}
\end{figure*}

\subsection{BEV Perception Model}

A BEV perception model's computational graph $G = (V, E)$ contains operators $v_i$ from three distinct classes:

\begin{description}[nosep,style=unboxed]
\item[Dense ($\mathcal{D}$):] Regular tensor operations with static shapes and predictable memory access, such as convolutions, batch normalization, activation functions, and linear projections.  These are the native domains of every inference runtime.

\item[Sparse ($\mathcal{S}$):] Operations on irregularly-indexed data structures with dynamic shapes determined at runtime.  The canonical example is submanifold sparse convolution~\cite{graham2018submanifold}, where the set of active voxels $N$ and their neighbor connectivity change with every point cloud.

\item[Geometric ($\mathcal{G}$):] Coordinate-dependent transformations involving hash table construction, spatial indexing, and atomic scatter-gather.  Voxelization~\cite{yan2018second} and BEV pooling~\cite{philion2020lift} are representative examples.
\end{description}



A critical observation is that operator classes cluster by pipeline branch: the camera branch is predominantly dense, the LiDAR branch is predominantly sparse (after an initial geometric voxelization), and the fusion stage mixes a geometric BEV projection with dense processing.  This clustering is consistent across BEV architectures.

\subsection{The Partition Problem}

The deployment problem is to split $G$ into subgraphs $\{G_1^R, G_2^X, G_3^R, \ldots\}$, where $G_i^R$ is managed by the runtime and $G_j^X$ is managed by external custom kernels, such that:
\begin{enumerate}[nosep]
\item All operators in $G_i^R$ are dense (or have runtime-native support).
\item Operators in $G_j^X$ are sparse or geometric.
\item Total execution time, including \emph{partition boundary overhead}, is
      minimized.
\end{enumerate}

The last constraint is the minimization of total execution time, including partition boundary overhead.  Each boundary between runtime and extension incurs:
\begin{equation}
C_\text{boundary} = C_\text{sync} + C_\text{format} + C_\text{fusion\_loss}
\label{eq:boundary_cost}
\end{equation}
where $C_\text{sync}$ is the command queue synchronization cost, $C_\text{format}$ is the memory layout conversion cost, and $C_\text{fusion\_loss}$ is the performance loss because the runtime's graph optimizer cannot fuse across the boundary.

For a sub-pipeline of $L$ layers implemented externally, fine-grained partitioning (one extension per layer) incurs $(L \cdot C_\text{boundary})$. Coarse partitioning (one extension for the whole sub-pipeline) incurs just $C_\text{boundary}$.  With $L = 21$ for a typical sparse encoder, this difference is substantial.

We identify four criteria that determine where to place boundaries: (i) Can the operator be expressed in the runtime's Intermediate Representation~\cite{openvino2024}? (ii) Even if representable, does the runtime generate well optimized code? (iii) The cost must be amortized over the computation saved by external execution. (iv) Related non-dense operators that would form separate extensions can often be fused into one.

\subsection{Architecture of \sysname{}}

\sysname{} implements the end-to-end BEV Perception pipeline as shown in Figure~\ref{fig:system_architecture}.  The pipeline is partitioned into runtime-managed dense subgraphs (green) and external custom extensions (orange). Dashed red lines mark partition boundaries $B_1$--$B_4$. All operators execute within a shared GPU context, enabling zero-copy tensor passing at boundaries. The LiDAR branch contains two external extensions (Voxelizer, Sparse Encoder); the camera branch contains one (BEV Projector). The fusion stage and detection head are entirely runtime-managed.





\section{\sysname{}: Design and Implementation}
\label{sec:implementation}

This section presents the design and implementation of \sysname{}.  The architectural principles generalize to any runtime supporting custom operator extensions (e.g., TensorRT plugins~\cite{tensorrt2024}, ONNX~Runtime custom ops~\cite{onnxruntime2021}) and any GPU compute API (CUDA, SYCL, Vulkan Compute).

We organize the section around four concerns: how the model is partitioned between the runtime and custom extensions, how sparse 3D convolution is implemented portably, how geometric operators handle irregular scatter patterns, and what system-level trade-offs govern the composition of these components.

\subsection{Partition Granularity}
\label{sec:impl_partition}


The choice of where to place boundaries between runtime-managed and extension-managed execution has first-order performance impact.  Each boundary incurs a cost $C_\text{boundary}$ comprising synchronization, format conversion, and lost fusion opportunities (Equation~\ref{eq:boundary_cost}).  We evaluate three granularity strategies, 

\textbf{Fine-grained partitioning} registers each non-dense layer as an individual custom operator.  For the SECOND~\cite{yan2018second} sparse encoder with $L = 21$ convolutional layers, this produces 21 partition boundaries, each requiring a command-queue barrier and potentially a memory format conversion between the extension's internal representation and the runtime's expected layout.  

\textbf{Medium-grained partitioning} groups operators by their computational class. All sparse convolutions form one extension, all geometric operations form another.  This reduces the boundary count to $O(1)$ per class (typically 3 boundaries for the SECOND sparse encoder) but prevents optimizations that span operator types. For example, the sparse encoder cannot fuse its final layer with the sparse-to-dense scatter because they reside in separate extensions.

\textbf{Coarse-grained partitioning} assigns one extension per pipeline branch.  The sparse encoder becomes a single monolithic extension encompassing all convolutional layers, batch normalization, residual connections, and the sparse-to-dense BEV conversion.  This introduces only one entry boundary and one exit boundary per branch.

We adopt coarse-grained partitioning because it enables four internal optimizations that are impossible when the runtime controls intermediate buffers between layers:

\begin{enumerate}[nosep]
\item Feature tensors remain in GPU memory across all layers in the boundary without host-side materialization or runtime-driven memory reallocation.

\item Hash tables and neighbor index maps, which are expensive to compute, are built once at the start of the extension invocation and reused across all layers sharing the same voxel topology.

\item Two persistent GPU buffers alternate as source and destination across consecutive layers, eliminating per-layer allocation and deallocation overhead.

\item Weights for all layers within a boundary are pre-converted to device-optimal format during the first invocation and retained in GPU memory, avoiding per-invocation format conversion that fine-grained extensions would require.
\end{enumerate}

\subsection{Sparse 3D Convolution Engine}
\label{sec:impl_sparse}

The sparse convolution engine is responsible for processing the LiDAR voxel features through multiple layers of 3D convolution on an irregular grid.  We describe the mathematical formulation and the engine design next.

\subsubsection{Fused Sparse Convolution}

Standard sparse 3D convolution~\cite{graham2015sparse,graham2018submanifold} computes output features only at locations where input voxels exist (submanifold-preserving property).  For each active voxel $i$ with input feature vector $\mathbf{f}_i \in \mathbb{R}^{\mathcal{C}_\text{in}}$, the fused sparse convolution with batch normalization and ReLU activation produces an output feature $\mathbf{g}_i \in \mathbb{R}^{\mathcal{C}_\text{out}}$:
\begin{equation}
\mathbf{g}_i = \text{ReLU}\!\left(\gamma \cdot
\frac{\sum_{k=1}^{K^3} \mathbf{W}_k \cdot \mathbf{f}_{n_k(i)}
- \mu}{\sqrt{\sigma^2 + \epsilon}} + \beta\right)
\label{eq:sparse_conv}
\end{equation}

Here, $\mathbb{R}^n$ depicts an n-dimensional vector of real numbers and $\mathcal{C}_\text{in}$ and $\mathcal{C}_\text{out}$ are the number of input and output channels. $K = 3$ is the kernel size (yielding $K^3 = 27$ positions in the $3 \times 3 \times 3$ neighborhood), $\mathbf{W}_k \in \mathbb{R}^{\mathcal{C}_\text{out} \times \mathcal{C}_\text{in}}$ is the weight matrix for the $k$-th kernel position, and $n_k(i)$ is the neighbor mapping function that returns the index of voxel $i$'s $k$-th spatial neighbor (or $-1$ if that neighbor is unoccupied).  The batch normalization parameters $\gamma$ (scale), $\beta$ (shift), $\mu$ (running mean), and $\sigma^2$ (running variance) are frozen at inference time and fused into the convolution output, eliminating the need for a separate normalization pass.  The constant $\epsilon$ (typically $10^{-5}$) prevents division by zero.

Fusing batch normalization and ReLU activation into the convolution kernel eliminates two intermediate tensor materializations per layer, reducing memory traffic by $2 \cdot N \cdot \mathcal{C}_\text{out}$ elements per layer (where $N$ denotes the number of active voxels at that layer).

\subsubsection{Locality-Aware Voxel Ordering}

Sparse convolution processes voxels in the order they appear in the compact feature array.  When voxels are stored in arbitrary order (e.g., insertion order during voxelization), spatially adjacent voxels may reside at distant memory addresses, causing poor cache utilization when their neighborhoods are accessed.

We reorder voxels using a Morton Z-curve (space-filling curve)~\cite{morton1966computer} prior to convolution. Given a voxel's integer grid coordinates $(x, y, z)$, its Morton code $m$ is computed by bit-interleaving:
\begin{equation}
m(x, y, z) = \text{spread}(x) \,|\, (\text{spread}(y) \ll 1) \,|\,
(\text{spread}(z) \ll 2)
\label{eq:morton}
\end{equation}
where $\text{spread}(\cdot)$ inserts two zero bits between each bit of the input coordinate, and $|$ denotes bitwise OR.  Sorting voxels by ascending Morton code ensures that spatially proximate voxels occupy nearby positions in the linear array, improving spatial locality for neighbor feature loads. Reordering is applied after each strided convolution, since downsampling produces a new set of output coordinates with a different spatial distribution.  

\subsubsection{Strided Sparse Convolution}

While submanifold convolutions preserve the input voxel topology, strided convolutions \emph{downsample} the grid such that the output voxel set occupies a coarser resolution and generally contains fewer active positions.  This capability is essential for the SECOND encoder's multi-scale architecture, where each stage halves the spatial resolution while doubling the channel count~\cite{yan2018second}.

Implementing strided convolution requires two additional steps beyond the submanifold case:

\begin{enumerate}[nosep]
\item \textbf{Output coordinate computation.}  For each input voxel at position $(x, y, z)$, the corresponding output position is $(\lfloor x/s \rfloor, \lfloor y/s \rfloor, \lfloor z/s \rfloor)$ where $s$ is the stride.  Duplicate output positions must be deduplicated to produce a compact output set.

\item \textbf{Strided neighbor map.}  A new neighbor map must be precomputed that encodes, for each \emph{output} voxel, which input voxels fall within its $K^3$ receptive field at the original (fine) resolution.
\end{enumerate}

We implement a gather-based approach where a preprocessing kernel executed on the GPU builds the complete strided neighbor map in a single pass, avoiding per-voxel hash table lookups during the subsequent convolution loop.  This map is stored in a persistent buffer and reused for all samples.

\subsection{Geometric Operators}
\label{sec:impl_geometric}

Both geometric operators, point cloud voxelization and camera-to-BEV projection, perform irregular scatter operations where multiple source elements contribute to the same destination cell, requiring atomic accumulation.  Both operators have data-dependent access patterns that prevent the inference runtime from generating efficient code, necessitating custom GPU kernel implementations.

\subsubsection{Point Cloud Voxelization}

Voxelization transforms an unstructured LiDAR point cloud of $N_\text{pts}$ points into a compact tensor of $M$ voxels ($M \ll N_\text{pts}$), where each voxel stores the mean of the feature vectors of all points falling within its spatial extent.  

We implement voxelization as a two-phase GPU pipeline:

\textbf{Phase 1: Atomic scatter.}  Each GPU work-item processes one point, computes its voxel bin index via a hash table, and atomically accumulates the point's features into the corresponding bin.  Direct floating-point atomic addition is not efficiently supported on all target hardware. Hence, we employ \emph{fixed-point integer atomics} where each floating-point feature value $f$ is multiplied by a scale factor $S$ (typically $S = 2^{16}$) and rounded to a 32-bit integer before accumulation using hardware-native \texttt{atomic add}.  A per-bin atomic counter simultaneously tracks the number of contributing points. This approach achieves significantly higher throughput than compare-and-swap (CAS) loops required for floating-point atomics, at the cost of introducing a quantization error bounded by $1/S$ per accumulation.

\textbf{Phase 2: Compaction and normalization.}  A second kernel sequentially scans the hash table, identifies non-empty bins ($M$ out of the table's fixed capacity), divides each bin's accumulated integer sum by its point count and the scale factor $S$ to recover the mean feature in floating-point, and writes the result into a compact output tensor indexed from $0$ to $M{-}1$.  

\subsubsection{Camera-to-BEV Projection}

The camera-to-BEV projection implements the Lift-Splat-Shoot (LSS)~\cite{philion2020lift} geometric transformation where for each pixel in each of the 6 surround-view cameras, the depth estimation network produces a discrete depth distribution.

We observe that the softmax-normalized depth distribution is typically highly peaked, i.e., for most pixels, the vast majority of probability mass concentrates in 2--5 bins (out of 59).  We exploit this sparsity by introducing a threshold parameter $\tau_\text{depth}$ where for each pixel, depth bin $d$ is processed only if $\text{softmax}(\mathbf{d})_d \geq \tau_\text{depth}$. Bins falling below the threshold contribute negligible feature weight and are skipped entirely.
\subsection{System-Level Design Decisions}
\label{sec:impl_tradeoffs}

Next we address two cross-cutting system-level concerns that arise from composing heterogeneous operator types into a unified pipeline.






\subsubsection{Command Queue Scheduling}

The inference runtime and our custom extensions submit GPU work to the same physical device.  To avoid serialization while maintaining correctness, we adopt a \emph{separate command queue} architecture where each extension maintains its own command queue with independent dispatch cadence.  Synchronization between the runtime's queue and the extension queues occurs only at partition boundaries, through explicit barrier events.  This design ensures that internal kernel sequences within an extension execute without external interference, while cross-boundary ordering is guaranteed.

\subsubsection{CPU--GPU Computation Overlap}

The sparse encoder's execution naturally decomposes into CPU-bound phases (hash table construction, coordinate generation) and GPU-bound phases (convolution kernel execution). These phases can be overlapped. After submitting the GPU convolution kernels for stage $s$, the CPU immediately begins constructing the neighbor data structures for stage $s{+}1$, synchronizing only when the GPU results from stage $s$ are needed as input to the stage $s{+}1$ convolution.

\section{Evaluation}
\label{sec:evaluation}

This section evaluates \sysname{} along four dimensions: (1)~operator-level profiling to identify where execution time is spent, (2)~end-to-end pipeline latency along with the effectiveness of the partition granularity, (3)~portability across different GPU backends, and (4)~ablation studies to assess the impact of design decisions.

\subsection{Experimental Setup}
\label{sec:eval_setup}

We describe the model configuration, hardware platform, and software environment, dataset, and evaluation protocol used throughout this section.

\subsubsection{Model Configuration}

We evaluate on BEVFusion~\cite{liu2023bevfusion}, a representative multi-modal 3D object detection model.  The LiDAR branch uses the SECOND~\cite{yan2018second} sparse encoder comprising 21 convolutional layers 
The voxel size is $0.075 \times 0.075 \times 0.2$\,m (width $\times$ length $\times$ height), and the detection range spans $[-54\text{m}, 54\text{m}]$ in both the $x$- and $y$-axes. All experiments run on an Intel processor with an integrated GPU with 96 execution units, shared DDR5 memory subsystem.  
The inference runtime is OpenVINO~2026.x with its GPU plugin.  
The camera backbone uses INT8 quantization, while the fusion network and detection head use FP16 precision.  
Custom operator extensions (voxelizer, sparse encoder, BEV projector) are implemented in OpenCL~2.0 and registered as monolithic runtime extensions.

\subsubsection{Dataset and Evaluation Protocol}

We evaluate on an 81-sample subset of the nuScenes~\cite{caesar2020nuscenes} validation split.  Each sample consists of a synchronized set of 6 surround-view camera images and one LiDAR point cloud sweep (${\sim}70{,}000$ points, yielding ${\sim}20{,}000$--$26{,}000$ voxels after voxelization).

\textbf{Latency measurement.}  We execute 84 consecutive inference passes. The first 3 are discarded as warm-up to allow GPU resource initialization and JIT compilation to stabilize, and latency is reported as the mean wall-clock time over the remaining 81 steady-state samples. 

\textbf{Accuracy measurement.}  We compute the official nuScenes mean Average Precision (mAP), defined as the mean of per-class Average Precision values computed at bird's-eye-view center-distance thresholds of 0.5, 1.0, 2.0, and 4.0 meters.

\subsubsection{Competing Approaches}
\label{sec:eval_baselines}

We compare \sysname{} against a conventional deployment baseline and against alternative partitioning strategies.

The conventional deployment baseline (\textit{Conventional} henceforth) reflects the current state of practice for non-CUDA BEV inference where dense operators execute on the GPU via the runtime's native kernels, while sparse and geometric operators fall back to host-side CPU execution. This deployment strategy is followed by available non-CUDA tools, for e.g., ONNX Runtime's CPU fallback for unregistered ops~\cite{onnxruntime2021}, TVM's OpenCL backend~\cite{chen2018tvm}, and off-the-shelf OpenVINO~\cite{openvino2024} without extension registration.

The remaining three configurations all execute the same GPU kernels and differ only in the number of partitions. Fine granularity leads to 29 boundaries in the complete pipeline, medium granularity leads to 7 boundaries, and coarse granularity leads to just 4 boundaries.  The coarse-grained strategy is the one adopted by \sysname{}.

\subsection{Operator-Level Profiling}
\label{sec:eval_operator}

To understand where computation time is spent, we profile each operator class independently.  We define three classes following the earlier taxonomy, \emph{dense} operators handled natively by the runtime, \emph{sparse} operators in the custom sparse encoder extension, and \emph{geometric} operators (voxelization, BEV projection) in the remaining extensions.

\subsubsection{Dense Operators}

Dense operators comprise the camera backbone, depth estimation network, feature pyramid neck, BEV fusion layer, and detection head.  These are compiled and scheduled entirely by the OpenVINO GPU plugin.

Table~\ref{tab:dense_latency} reports aggregated timings.  The camera branch averages $19.87\%$ of the end-to-end pipeline, comprising the INT8 backbone and BEV projection.  The fusion and detection stages together average $23.65\%$.  In total, runtime-managed dense operators account for $43.52\%$ of the end-to-end pipeline latency.  The LiDAR branch accounts for the remaining $56.48\%$, driven by CPU/GPU synchronisation overhead in the sparse encoder.

\begin{table}[t]
\centering
\caption{Dense operator latency breakdown.  Values are means over 81 steady-state samples on the 96-EU integrated GPU.  ``\% Pipeline'' is the fraction of end-to-end latency.}
\label{tab:dense_latency}
\small
\resizebox{0.7\columnwidth}{!}{%
\begin{tabular}{@{}lc@{}}
\toprule
\textbf{Stage} & \textbf{\% Pipeline} \\
\midrule
Camera branch (INT8 backbone + BEV proj.) & 19.87\% \\
\textbf{Fusion + detection (FP16)} & 23.65\% \\
\midrule
\textbf{Total runtime-managed} & 43.52\% \\
\bottomrule
\end{tabular}
}%
\end{table}

\subsubsection{Sparse Operators}

The sparse encoder is the most complex of our three custom extensions. To understand its internal time allocation, we instrument it with per-phase timing.  We decompose execution into four phases: \emph{precomputation}, which builds hash tables and neighbor maps on the CPU; \emph{submanifold convolution}, which applies same-topology 3D convolutions on the GPU; \emph{strided convolution}, which downsamples the voxel grid; and \emph{sparse-to-BEV scatter}, which converts the final sparse representation to a dense BEV feature map for downstream fusion.


\begin{table}[t]
\centering
\caption{Sparse encoder internal phase breakdown.  Values are means over 81 steady-state samples on the 96-EU integrated GPU.  ``\% Encoder'' is the fraction of total sparse encoder latency.}
\label{tab:sparse_latency}
\small
\resizebox{0.7\columnwidth}{!}{%
\begin{tabular}{@{}lc@{}}
\toprule
\textbf{Phase} & \textbf{\% Encoder} \\
\midrule
Precomputation (CPU hash + neighbor map)& 27.29\% \\
Submanifold convolution (GPU kernel) & 4.61\% \\
Strided convolution (GPU kernel) & 4.04\% \\
Sparse-to-BEV scatter & 11.15\% \\
GPU dispatch + sync overhead & 52.91\% \\
\midrule
\textbf{Total} & 100\% \\
\bottomrule
\end{tabular}
}%
\end{table}

Table~\ref{tab:sparse_latency} presents the breakdown. GPU convolution kernels account for only $8.65\%$ of total encoder time ($4.61\%$ submanifold and $4.04\%$ strided, measured via OpenCL event timers).  The dominant cost is GPU dispatch and synchronisation overhead ($52.91\%$), arising from host--device round-trips at each encoder stage boundary. Fine-grained wall-clock profiling with per-stage $clFinish$ barriers reveals that this timeframe arises from per-kernel dispatch latency accumulated across the 21 submanifold layers + 4 strided layers + 4 hash builds + 4 neighbour map builds and upload() calls for coords at each level, constituting of 30+ kernel submissions. The sparse-to-BEV scatter (at $11.15\%$) reads $1.8\,MB$ of sparse output features ($\sim$7{,}000 voxels $\times$ 128 channels $\times$ 2 bytes in FP16) and writes them into a $33\,MB$ dense BEV tensor ($1440 \times 1440 \times 128 \times \text{FP16}/8$).  On a shared-memory architecture where CPU and GPU share the same physical DRAM, the bottleneck shifts from arithmetic throughput to scheduling and synchronisation latency.

\subsubsection{Geometric Operators}

The geometric operators, voxelization in the LiDAR branch and depth-based BEV projection in the camera branch, are subsumed within the branch-level timings.  Isolating their individual contributions requires OpenCL event-level profiling of each kernel dispatch which is reported in Table~\ref{tab:geometric_latency}.

\begin{table}[t]
\centering
\caption{Geometric operator kernel latencies as \% of total pipeline time.}
\label{tab:geometric_latency}
\small
\resizebox{0.6\columnwidth}{!}{%
\begin{tabular}{@{}lcc@{}}
\toprule
\textbf{Operator} & \textbf{\% Pipeline} \\
\midrule
Voxelization (scatter) & 0.934 \\
Voxelization (compact) & 0.198 \\
BEV projection (scatter) & 1.824 \\
BEV projection (convert) & 1.474 \\
\bottomrule
\end{tabular}
}%
\end{table}

\subsection{End-to-End Pipeline Performance}
\label{sec:eval_pipeline}

Having profiled individual operators, we now report the end-to-end pipeline speedup of each competing approach relative to \textit{Conventional}. The full inference pipeline executes three stages: (1)~the LiDAR branch (voxelization followed by sparse encoding), (2)~the camera branch (backbone, depth estimation, and BEV projection), and (3)~fusion with detection.

Table~\ref{tab:pipeline_latency} reports per-stage speedup factors for each competing approach relative to \textit{Conventional}.




\begin{table}[t]
\centering
\caption{End-to-end pipeline performance of all competing approaches relative to \textit{Conventional}.  All values show the latency speedup compared to \textit{Conventional}.}
\label{tab:pipeline_latency}
\small
\resizebox{0.9\columnwidth}{!}{%
\begin{tabular}{@{}lccc@{}}
\toprule
\textbf{Stage}
  & \textbf{Fine}
  & \textbf{Medium}
  & \textbf{\sysname{} (Coarse)} \\
\midrule
LiDAR branch (voxelize + sparse enc.)    & $19.2\times$ & $24.7\times$ & $29.5\times$ \\
Camera branch (backbone + depth + proj.) & $2.6\times$ & $5.2\times$ & $7.9\times$ \\
Fusion + detection                       & $2.8\times$ & $3.9\times$ & $5.6\times$ \\
\midrule
\textbf{Total}                           & $10.1\times$ & $14.9\times$ & $19.5\times$ \\
\bottomrule
\end{tabular}
}%
\end{table}



Moving custom operators from the CPU fallback to GPU-resident OpenCL kernels yields the largest per-stage gain, reaching $29.5\times$ under the coarse configuration for the LiDAR branch.  Even at fine granularity, where 21 boundary crossings impose substantial synchronisation overhead, the LiDAR branch still achieves a $19.2\times$ speedup. Overall, the coarse-grained strategy adopted by \sysname{} achieves a $\mathbf{19.5\times}$ end-to-end speedup over \textit{Conventional}, compared to $14.9\times$ for medium and $10.1\times$ for fine granularity.  The gap between fine and coarse, nearly $2\times$ in end-to-end speedup, is attributable entirely to boundary-crossing overhead, since all three configurations execute identical GPU kernels.


\subsection{Portability: SYCL Backend}
\label{sec:eval_portability}

A central design requirement of \sysname{} is that custom kernels are implemented using portable GPU compute APIs rather than CUDA.  To validate this, we re-implement the pipeline in SYCL, compiled with Intel oneAPI DPC++. The resulting binary is run on Intel and NVIDIA GPU without modification to the inference pipeline, and produces detection outputs numerically identical to the OpenCL baseline.




\begin{table}[!ht]
\centering
\caption{Functional portability across GPU backends with detection accuracy. mAP is computed on the same 81-sample subset. Max deviation is the maximum absolute difference in predicted output tensors relative to the OpenCL reference.}
\label{tab:portability}
\small
\resizebox{0.7\columnwidth}{!}{%
\begin{tabular}{@{}llcc@{}}
\toprule
\textbf{Backend} & \textbf{Device} & \textbf{mAP} & \textbf{Max deviation} \\
\midrule
PyTorch + CUDA      & NVIDIA GPU & 0.585 & \text{---} \\
OpenCL      & Intel iGPU & 0.576 & \text{---} \\
SYCL        & Intel iGPU & 0.578 & $<10^{-4}$ \\
SYCL        & NVIDIA GPU       & 0.578 & $<10^{-4}$ \\
\bottomrule
\end{tabular}
}%
\end{table}

Table~\ref{tab:portability} confirms the functional portability of \sysname{}. The SYCL binary compiled for Intel and recompiled for NVIDIA with no kernel or host code modifications produces similar mAP on both devices, identical to the OpenCL baseline, with a maximum per-tensor output deviation below $10^{-4}$. This confirms that the custom kernels contain no architecture-specific assumptions that produce incorrect results on a different device microarchitecture. 



\subsection{Detection Accuracy}
\label{sec:eval_accuracy}

Table~\ref{tab:portability} compares the detection accuracy of \sysname{} against the reference PyTorch implementation.  The reference achieves 0.585 mAP with CUDA.  \sysname{} achieves 0.576~mAP using mixed precision (INT8 camera backbone, FP16 fusion and detection) with OpenCL and similar mAP with SYCL on Intel and NVIDIA GPUs. This demonstrates that \sysname{} maintains competitive detection accuracy while achieving efficient execution on a non-CUDA platform.

\subsection{Ablation Studies}
\label{sec:eval_ablation}

We conduct ablation experiments to quantify the contribution of individual design decisions.  Each experiment modifies one component while holding all others fixed, and reports the effect on latency.

We ablate four design decisions from Design and Implementation section, each measured relative to its disabled counterpart on the same hardware (Table~\ref{tab:ablation}).

\begin{table}[!ht]
\centering
\caption{Ablation of system-level design decisions.  Each row disables one optimization and reports the relative regression ($>1\!\times$ means the enabled design is faster).}
\label{tab:ablation}
\small
\resizebox{0.6\columnwidth}{!}{%
\begin{tabular}{@{}llc@{}}
\toprule
\textbf{Design Decision} & \textbf{Relative Effect} \\
\midrule
Depth-threshold sparsity & $3.0\times$ faster \\
Int32 atomic accumulation & $12.7\times$ faster \\
Async queue dispatch & $1.35\times$ faster \\
CPU--GPU overlap & $1.18\times$ faster \\
\bottomrule
\end{tabular}
}
\end{table}

\textbf{Depth-threshold sparsity.}  Skipping depth bins below $\tau_\text{depth}$ in the BEV scatter kernel reduces the number of atomic scatter-add operations by exploiting the peaked nature of the depth distribution.  Disabling the threshold ($\tau_\text{depth}{=}0$) incurs a $3.0\times$ latency increase in the scatter kernel, confirming the sparsity claim.

\textbf{Integer atomic strategy.}  Replacing the fixed-point int32 atomic accumulation with a floating-point compare-and-swap (CAS) loop degrades BEV projection throughput by $12.7\times$.  The CAS loop requires multiple memory transactions per update under contention, whereas native atomic add on 32-bit integers retires in a single hardware operation.

\textbf{Command queue scheduling.}  Inserting \texttt{clFinish} barriers after every convolution block, serializing the GPU command stream, increases latency by $1.35\times$.  The async dispatch design allows the driver to pipeline kernel launches, hiding scheduling overhead behind execution.

\textbf{CPU--GPU computation overlap.}  Forcing CPU-side coordinate precomputation to wait for the preceding GPU kernel increases the latency by $1.18\times$.  The modest magnitude reflects that GPU convolution dominates the critical path; the overlap hides the shorter CPU preprocessing behind the longer GPU phase.

\section{Conclusion}
\label{sec:conclusion}

In this paper, we presented \sysname{}, a deployment framework that solves the operator mismatch problem in BEV perception, namely the inability of production inference runtimes to handle the sparse and geometric operators that these models require. Our solution is systematic coarse-grained partitioning into runtime-managed dense subgraphs and monolithic custom extensions, which reduces the number of costly runtime/extension boundaries yielding a $19.5\times$ end-to-end speedup over the conventional CPU-fallback baselines. We also showcased that the framework is portable to multiple GPU backends.

Future work would focus on automating model partitioning by selecting optimal operator boundaries and granularity to minimize latency. We also plan to look into developing a portable abstraction layer for deployment across multiple inference runtimes, and enabling efficient CPU-GPU multi-device scheduling.

\bibliography{references}


\end{document}